\documentclass{article}

\pdfoutput=1
    \PassOptionsToPackage{numbers}{natbib}

\usepackage{graphicx}
\usepackage{microtype}
\usepackage{graphicx}
\usepackage{subfigure}
\usepackage{booktabs} 
\usepackage{multirow}
\usepackage{hyperref}

\usepackage{amsmath}
\usepackage{amssymb}
\usepackage{mathtools}
\usepackage{amsthm}
\usepackage{nicematrix}
\usepackage{makecell}
\usepackage[capitalize,noabbrev]{cleveref}
\usepackage{hyphenat} 
\usepackage{algorithm}
\usepackage{algorithmic}

\theoremstyle{plain}
\newtheorem{theorem}{Theorem}[section]
\newtheorem{proposition}[theorem]{Proposition}

\theoremstyle{definition}
\newtheorem{definition}[theorem]{Definition}

\theoremstyle{remark}

\usepackage{hyperref}

\usepackage[preprint]{neurips_2023}

\usepackage[utf8]{inputenc} 
\usepackage[T1]{fontenc}    
\usepackage{url}            
\usepackage{booktabs}       
\usepackage{amsfonts}       
\usepackage{nicefrac}       
\usepackage{microtype}      
\usepackage{xcolor}         
\usepackage{natbib}
\usepackage{enumitem}

\title{Towards Optimal Compression: Joint Pruning and Quantization}

\author{%
  Ben Zandonati \\
  University of Cambridge\\
  \texttt{baz23@cam.ac.uk} \\
   \And
   Glenn Bucagu  \\
  University of Zürich \\
   \texttt{glennanta.bucagu@uzh.ch} \\
   \And
   Adrian Alan Pol \\
  Princeton University \\
   \texttt{ap6964@princeton.edu} \\
   \AND
   Maurizio Pierini \\
   CERN \\
   \texttt{maurizio.pierini@cern.ch} \\
   \And
   Olya Sirkin, Tal Kopetz\\
   CEVA. Inc \\
  \{\texttt{olya.sirkin, tal.kopetz}\}@ceva-dsp.com\\
}

\begin{document}

\maketitle

\begin{abstract}
  Model compression is instrumental in optimizing deep neural network inference on resource-constrained hardware. The prevailing methods for network compression, namely quantization and pruning, have been shown to enhance efficiency at the cost of performance. Determining the most effective quantization and pruning strategies for individual layers and parameters remains a challenging problem, often requiring computationally expensive and ad hoc numerical optimization techniques. This paper introduces FITCompress, a novel method integrating layer-wise mixed-precision quantization and unstructured pruning using a unified heuristic approach. 
  By leveraging the Fisher Information Metric and path planning through compression space, FITCompress optimally selects a combination of pruning mask and mixed-precision quantization configuration for a given pre-trained model and compression constraint. 
  Experiments on computer vision and natural language processing benchmarks demonstrate that our proposed approach achieves a superior compression-performance trade-off compared to existing state-of-the-art methods.
  FITCompress stands out for its principled derivation, making it versatile across tasks and network architectures, and represents a step towards achieving optimal compression for neural networks.
\end{abstract}

\section{Introduction}\label{sec:Introduction}

Deep neural networks (DNNs) require substantial computational resources for both training and deployment. However, many real-world applications involve devices with limited computational capabilities, such as mobile phones, embedded systems, or edge devices \cite{coelho2021automatic, 9283584, 9309177, zhang2020deep}. Consequently, the demand for {\em model compression}~\cite{sze2017efficient} methods has surged with the scaling of large computer vision and natural language processing (NLP) foundational models. These methods are typically designed to reduce model size, latency, energy or memory consumption, while minimizing performance loss.

{\em Pruning} and {\em quantization} methods address these issues by reducing the memory footprint and computational requirements of neural networks while maintaining acceptable performance levels.

Quantization reduces the precision of network parameters, typically from floating-point to fixed-point representation. By using fewer bits to represent each parameter, quantization significantly reduces the memory requirements and computational complexity of network operations. This is especially beneficial for hardware platforms that have limited memory and processing power such as FPGAs. Quantization can be performed homogeneously (i.e., using the same bit-width for the whole network), but empirical evidence suggests that each layer in a model exhibits a different sensitivity to quantization~\cite{dong2019hawq}. This calls for {\em mixed-precision quantization} (MPQ)~\cite{wu2018mixed}. The challenge, in this case, is the combinatorial explosion in the search space of possible MPQ configurations as networks grow. In particular, the time complexity is in $\mathcal{O}(\lvert B \rvert^{2L})$, where $B$ is the set of considered bit-widths and $L$ is the number of layers.

Pruning selectively removes unnecessary connections or parameters from the network, effectively reducing its size. It exploits the observation that many network parameters are redundant or contribute minimally to overall network performance. As with MPQ, as networks grow, the number of possible pruning masks to apply grows exponentially in $\mathcal{O}(2^{|\Theta|})$, where $|\Theta|$ is the number of parameters in the network. Previous works typically employ heuristic-based methods to determine parameter importance and prune accordingly. By removing such parameters, pruning reduces the model size, memory footprint, and computational requirements during inference.


Recent attempts have been made to optimally quantize and prune networks to improve the performance-compression trade-off for significant compression levels. Such methods typically employ slow, ad hoc, black-box or first-order optimization methods, the results of which are not guaranteed to generalize to future foundational models and challenging architectures. In particular, many methods sequentially prune then quantize ~\cite{han2016deep}, and do not consider the two under the same theoretical framework. 

In this work, we introduce FITCompress, a principled algorithm that combines MPQ and pruning for one-shot optimal network compression. By modelling MPQ and pruning under the Fisher Information Metric (FIM), we provide a method to integrate these two compression techniques. In doing so, we  leverage modern platforms that support sparse and quantized formats~\cite{DBLP:journals/corr/abs-2111-13445, pmlr-v119-kurtz20a, mishra2021accelerating}. This framework enables us to measure distances in the parameter space associated with compression. Moreover, we introduce a reduced action space for quantization and pruning, referred to as the \textit{compression space}. By traversing this space using simple path-planning techniques, we obtain compression configurations that satisfy the required resource constraints while minimizing the performance drop of a pre-trained model. FITCompress approximates perturbations in the parameter space and takes iterative steps towards achieving optimal compression.

Our main contributions are summarized as follows:

\begin{enumerate}[wide, labelindent=0pt]

    \item This paper introduces FITCompress, a principled algorithm that combines quantization and pruning under FIM to improve network compression. FITCompress provides a method to integrate these techniques, measures distances in the parameter space associated with compression, and traverses a reduced action space to find optimal compression configurations in one-shot.
    \item Empirically, we show state-of-the-art compression ratios across a diverse set of architectures and datasets. The architectures evaluated include LeNet-5, VGG-7, ResNet-18, ResNet-34, ResNet-50, BERT-Base, and YOLOv5. The datasets include MNIST, CIFAR-10, ImageNet, SST-2, MNLI-m, SQuAD and COCO. Via this extensive evaluation, we highlight the strong performance and generality of FITCompress.
 
\end{enumerate}

\section{Related Work}\label{sec:Related Work}

FITCompress employs layer-wise mixed-precision quantization and unstructured pruning. That is, quantization is performed at the granularity of layers, while pruning is performed at the granularity of individual parameters. Both are well supported in modern deep learning frameworks. Additionally, after quantization and pruning, further training is performed to recover lost performance. Such quantization-aware training (QAT) \cite{jacob2018quantization} is commonly employed by machine learning practitioners, and by all of our evaluation benchmarks (except where specified for YOLOv5 \cite{https://doi.org/10.5281/zenodo.7347926}). To that end, we present a brief taxonomy of relevant MPQ and pruning methods. For more comprehensive summaries of related work, we refer readers to the appropriate reviews~\cite{blalock2020state, hoefler2021sparsity, nagel2021white}.

\paragraph{Pruning.} 
Unstructured pruning involves removing individual model parameters according to a simple-to-compute heuristic. This can include the magnitude \cite{gale2019state, han2016deep, janowsky_1989}, first-order information \cite{lee2021layeradaptive, lee2018snip, molchanov2019importance, mozer1988skeletonization,  sanh2020movement, wang2020picking, zhang2022platon}, or second-order information \cite{frantar2021mfac, hassibi1992second, lecun1989optimal}. These methods suffer from {\em layer collapse}, in which all (or a significant number) of the weights in a single layer are pruned, rendering training and inference impossible. This issue is alleviated by conserving what \cite{tanaka2020pruning} refer to as synaptic saliency. While iterative pruning methods help to conserve synaptic saliency, better metrics (such as the Fisher Information \cite{ singh2020woodfisher, theis2018faster}) further boost pruning performance. Our work employs an approximation of FIM, iteratively computed as we move through the compression space. 

\paragraph{Mixed-Precision Quantization.} 
Second-order information is commonly used to effectively quantize neural networks. For example, in computer vision, \cite{dong2020hawq, dong2019hawq, yao2021hawq} used the Hessian's top eigenvalues/trace to determine a layer’s sensitivity to the perturbations associated with a reduction in bit precision. For NLP tasks, \cite{shen2020q} defined sensitivity measurement based on both the mean and variance of the top Hessian eigenvalues to generate MPQ configurations. Other black-box approaches have been proposed, including multi-stage optimization \cite{pandey2023practical, schaefer2023mixed}, linear programming \cite{ma2022ompq, tang2023mixedprecision}, deep reinforcement learning (DRL) \cite{elthakeb2020releq, qian2020channelwise, wang2019haq}, and neural architecture search (NAS) \cite{guo2020single, wu2018mixed} however these methods are computationally expensive. Recently, the Fisher Information Trace (FIT) \cite{zandonati2022fit} emerged as the superior metric for model sensitivity to quantization, being computationally inexpensive and correlating highly with final model performance. This work uses an extension of FIT, to cope with the significant deviations in the parameter space.

\paragraph{Joint Pruning and Quantization.} Combining pruning and quantization is nontrivial. Bayesian and variational approaches are commonly used \cite{louizos2017bayesian, tung_mori_2018, van2020bayesian, wang2020differentiable} but are difficult to scale to larger models. Hessian-based methods have been proposed \cite{frantar2022optimal}, but suffer from slow computation. More recently, \cite{hu2022opq} used non-linear constrained optimisation to analytically solve the joint MPQ and pruning problem. Notably, the method requires two fine-tuning stages. Similar multi-stage frameworks that prune then quantize were also introduced by \cite{han2016deep} and \cite{yu2020joint}. However, \cite{zhang2021training} empirically demonstrated the non-commutativity of pruning and quantization across generative and discriminatory tasks. This is one of the primary motivations for the path-planning approach in FITCompress. In summary, FITCompress differentiates itself by unifying and interleaving quantization and pruning under the computationally inexpensive Fisher Information.

\section{Methodology}\label{sec:Methodology}

Consider a dataset $D = \{(x, y)_{i}\}_{1}^{N}$ drawn i.i.d. from the true underlying joint distribution of samples. With this dataset, a parameterized model is trained to estimate the conditional distribution $p(y | x, \theta) = p_{\theta}$, where the parameters $\theta \in \Theta$ are chosen to minimize the loss function $\mathcal{L}_{\theta}(x, y): \mathcal{X} \times \mathcal{Y} \rightarrow \mathbb{R}$ via gradient descent. The parameterized model $p_{\theta}$ is a DNN consisting of $L$ layers. 

In this paper, we consider both layer-wise MPQ and unstructured pruning. Consequently, compression involves choosing each layer's bit-width $\{Q_{i}\}_{1}^{L}$ and providing a pruning mask $P \in \{0, 1\}^{|\Theta|}$. We will describe pruning in terms of the induced compression $\kappa =\frac{\sum{P}}{|\Theta|}$, where $\sum{P}$ is the number of masked parameters, and $|\Theta|$ is the total number of parameters in the model. Each layer-wise quantization and pruning mask defines a configuration $c \in \mathcal{C}^{L + |\Theta|}$ in the \textit{compression space}.

As noted in Section \ref{sec:Related Work}, searching this space, such that model performance is retained after compression, is the subject of ongoing research. In particular, principled methods derived from the Fisher Information~\cite{theis2018faster,zandonati2022fit} yield high-performing models at low compression rates. 

\subsection{FIT: Metric for Parameter Sensitivity}

Consider a general perturbation to the model parameters: $p_{\theta + \delta \theta}$, which could arise as a result of quantization or pruning. To measure the effect of this perturbation on the model, we use the KL divergence: $$D_{KL}(p_{\theta}||p_{\theta+\delta\theta})= \frac{1}{2} \delta\theta^{T} I(\theta) \delta\theta$$ Where the FIM~\cite{amari2016information}, $I(\theta)$, takes the following form (for a typical cross-entropy loss function): 
\begin{equation}
    I(\theta) = \mathbb{E} (\nabla_{\theta}\log p_\theta\nabla_{\theta}{\log p_\theta}^{T})
\end{equation}
As noted in Section \ref{sec:Related Work}, recent work \cite{zandonati2022fit} numerically approximates this distance in the parameter space associated with quantization. The resulting FIT metric proved an effective, and computationally inexpensive, quantization heuristic (shown here in empirical form):
\begin{equation}
    FIT(\theta, \delta\theta) = \sum_{l}^{L} \frac{1}{n(l)} \cdot Tr(\hat{I}(\theta_{l})) \cdot ||\delta\theta||_{l}^{2}
\end{equation}
Where $n(l)$ denotes the specific number of parameters in a single layer $l$, and $Tr(\hat{I}(\theta_{l}))$ denotes the trace of the empirical Fisher $\hat{I}$ for the set of parameters $\theta_{l}$ in a single layer $l$.

From an information geometric perspective, it is well known~\cite{amari2016information, nielsen2020elementary} that this formulation relies on a first-order divergence approximation, valid for small perturbations. It is clear that at very low bit precision and high pruning rates (that is to say, significant compression), this approximation is broken. More precisely, we move too far from the original model in the tangent space of our statistical manifold (induced by $p_{\theta}$). For a more detailed derivation, see Appendix \ref{sec: further_alg_details}.



\subsection{Pruning with FIT}


The consequences of beyond-small approximations have been indirectly addressed by the pruning literature. Recent work~\cite{lee2018snip} highlights the importance of the pruning criterion itself and~\cite{tanaka2020pruning} shows that iterative pruning methods improve upon single-shot schemes.

\begin{figure}[!ht]
    \centering
    \includegraphics[width=0.7\linewidth]{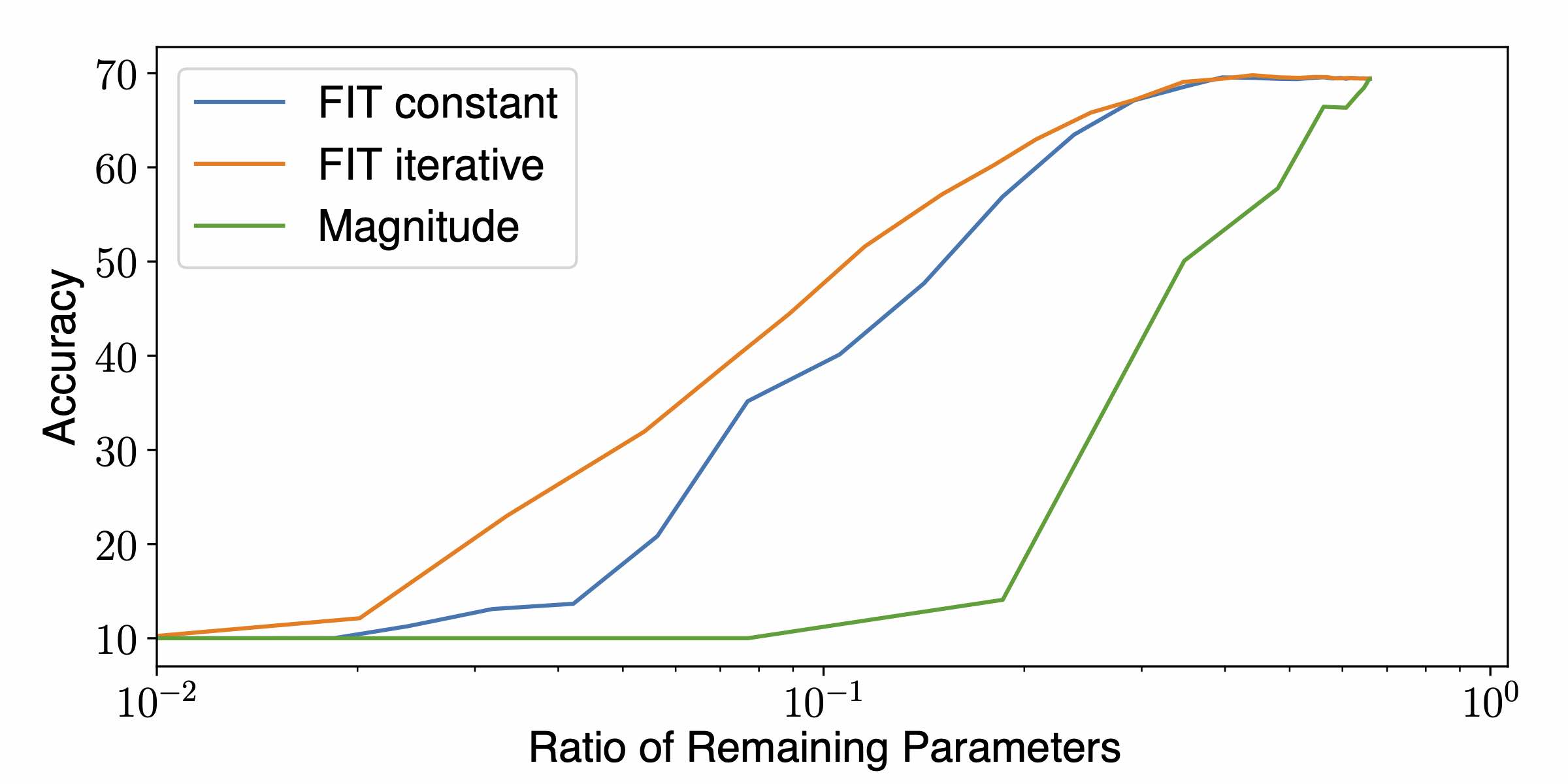}
    \caption{Fraction of non-zero parameters, $\kappa$, and accuracy after each iteration of pruning LeNet-5 on CIFAR-10 dataset. The lower the $\kappa$, the higher the compression ratio.}
    \label{fig:fit_pruning}
\end{figure}

From a geometric perspective, standard magnitude pruning assumes that the local metric tensor is an identity function. Similarly, non-iterative methods assume that the local tangent space is sufficiently representative of the statistical manifold, such that the metric remains constant. As shown in Figure~\ref{fig:fit_pruning}, removing both assumptions in the example of LeNet-5 on CIFAR-10 improves empirical performance. This repeats the findings of~\cite{tanaka2020pruning}. 

\subsection{Scheduling through Path Planning}

Equipping algorithms with a distance to explore compression space effectively is of central importance. When moving beyond local perturbations, the Fisher-Rao distance is obtained~\cite{atkinson1981rao}:

\begin{equation}
g = \int [\delta\theta^{T}I(\theta)\delta\theta]^{\frac{1}{2}}
\end{equation}
From this perspective, we define \textit{optimal compression}.

\begin{definition}(Optimal Compression) \label{def:optimal_compression}
An optimal compression configuration $c^{*}$ for a given constraint $\alpha$ is the solution to the following:

\begin{equation}
c^{*} = \arg\min_{c}\left(\int_{S(\theta, \theta_{c})} [\delta\theta^{T}I(\theta)\delta\theta]^{\frac{1}{2}}\right)
\end{equation}
Where $S(\theta, \theta_{c})$ refers to the geodesic (shortest) path, and $\theta_{c}$ obeys the compression constraint, i.e. $\alpha_c \le \alpha$.

\end{definition}

Direct use of Definition \ref{def:optimal_compression} is not possible. Compression space is discrete, and the evaluation of this integral is not computationally feasible. Both the final compression configuration, as well as the shortest path $S(\theta, \theta_{c})$ are unknown. Instead, we look to approximations of Definition \ref{def:optimal_compression}.

First, reducing the dimensionality of this compression space ($L+|\Theta|$) is necessary. We construct a reduced action space, where at each iteration, a layer's bit precision $Q_{l}$ may be reduced, or $\kappa$ is increased (according to specified schedules). In this case, quantization and pruning are reduced to path planning through the compression space. This process is shown in Figure \ref{fig:reduced_space}, and empirical examples are discussed in Section \ref{sec:Experiments}.

\begin{figure}[!ht]
\centering
\resizebox{0.8\linewidth}{!}{%
\begin{tabular}{ccccccc}
 &        $c_{0}$           &       $a_{0}$            &        $c_{1}$           &      $a_{1}$             &      $c_{2}$    &     $c_{i}$     \\
  $Q_{0}$& \multirow{4}{*}{$\begin{bmatrix}8\\8\\8\\8\\0\end{bmatrix}$} & \multirow{4}{*}{$\begin{matrix}\\\\\\\\\rightarrow\end{matrix}$} & \multirow{4}{*}{$\begin{bmatrix}8\\8\\8\\8\\0.3\end{bmatrix}$} & \multirow{4}{*}{$\begin{matrix}\\\\\rightarrow\\\\\\\end{matrix}$} & \multirow{4}{*}{$\begin{bmatrix}8\\8\\4\\8\\0.3\end{bmatrix}$} & \multirow{4}{*}{\dots\quad$\begin{bmatrix}4\\3\\4\\4\\0.97\end{bmatrix}$\quad\dots} \\
  $Q_{1}$&                   &                   &                   &                   &           &        \\
  $Q_{2}$&                   &                   &                   &                   &            &       \\
  $Q_{3}$&                   &                   &                   &                   &             &     \\
  $\kappa$&                   &                   &                   &                   &           &       
\end{tabular}}
\caption{Example of the reduced action space for compression. FITCompress takes actions $a_i$ to compress the model, either via quantizing a specific layer, or pruning.}
\label{fig:reduced_space}
\end{figure}

Traditional algorithms for solving these problems, such as Monte Carlo Tree Search or variations of DRL, typically do not generalize well to arbitrary networks, architectures, and schedules. Additionally, training times would be a prohibitive factor, as each trajectory requires additional fine-tuning after a compression configuration is generated. In lieu of these, we provide a practical and general algorithm, FITCompress, to generate candidate compression configurations. 

\subsection{FITCompress as Informed Search}

First, the distance $g$ is discretized, and FIT is used for numerical approximation:
\begin{equation}
\hat{g} = \sum_{a_i}FIT(\theta_{i}, \delta\theta_{i})^{\frac{1}{2}}
\end{equation}
Where the set of actions $\{a_i\}$ defines both the path through the compression space, as well as the final compression configuration. However, choosing greedily, to minimize each candidate model's contribution to  $\hat{g}$, does not provide optimal solutions. More precisely, this results in improper scheduling of quantization and pruning steps, and hence, sub-optimal configurations. 

To alleviate this issue, inspired by the A* search algorithm and other informed search algorithms \cite{DBLP:books/aw/RN2020}, we construct two search heuristics. One defines the distance of the path from the starting model to the candidate model $\hat{g}$ (as above), and the other estimates the minimum distance between the candidate model and the final compressed model $\hat{f}$. The latter is more challenging to define, as we do not know the final configuration, only the compression constraints to which $\theta_{c}$ must adhere. Instead, we choose to uniformly apply these constraints and use the current model to estimate the final distance: 
\begin{equation}
\hat{f} = |\alpha_{i} - \alpha|FIT(\theta_{i}, \theta_{i})^{\frac{1}{2}} 
\end{equation}
Where $|\alpha_{i} - \alpha|$ denotes the difference between the current compression and the required constraint. As we move closer to the final compression configuration, the empirical FIT estimation improves, as does $\hat{f}$. However, this remains a large approximation, and as such, we include an additional balancing parameter $\lambda$ such that actions are chosen greedily w.r.t. $\hat{g} + \lambda\hat{f}$ (chosen to be $0.5$).

 If $\hat{f}$ is not included, path planning fails to balance quantization and pruning, leading to sub-optimal compression configurations. We include pseudocode for FITCompress implementation in Algorithm~\ref{alg:fitcompress}.

\begin{figure}[!ht]
    \centering
    \includegraphics[scale=0.2]{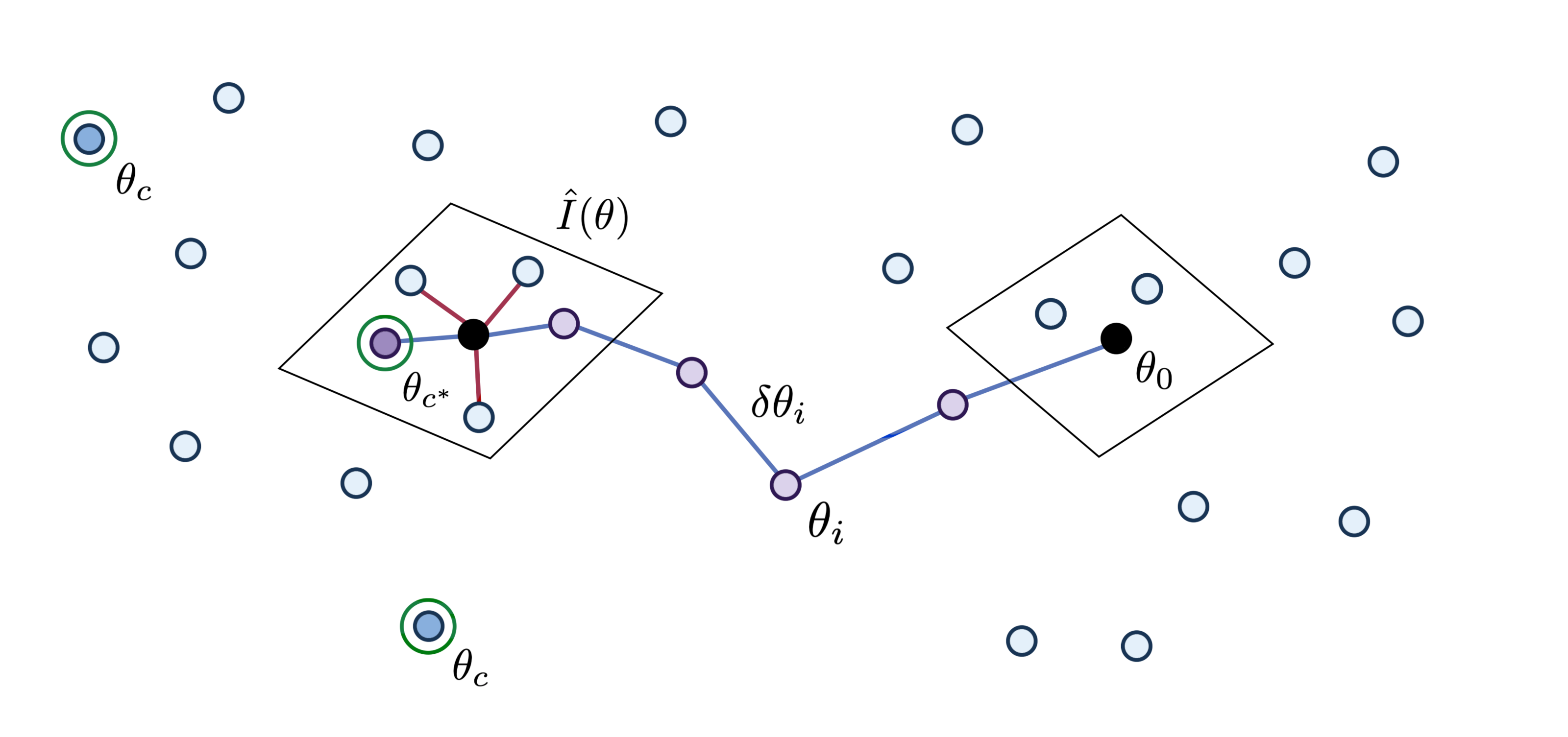}
    \caption{Illustration of the FITCompress algorithm, path planning through compression space. $\theta_0$ is the starting parameter set. $\theta_c$ denotes hypothetical configuration and $\theta_{c^*}$ is the optimal one. Using a single contribution to $\hat{g}$ for complete compression can become a false compass, as the optimal intermediate steps are not considered. Compressing the model directly could lead the algorithm to $\theta_{c}$ instead of $\theta_{c^*}$. Thus, inclusion of path planning is one of our central contributions.}
    \label{fig:path_planning}
\end{figure}

\begin{algorithm}[tb]
   \caption{FITCompress: Single-Shot Compression}
   \label{alg:fitcompress}
\begin{algorithmic}
   \REQUIRE Pre-trained model $p_{\theta_{0}}$, training dataset $D$, Quantization Schedule $Q$, Pruning Schedule $\kappa$, Desired Compression Constraint $\alpha$.
   \STATE {\bfseries Input:} initial parameters $\theta_{0}$, distance $\hat{g}_{0} = 0$, initial configuration $c_i = [32, 32, ..., 32, 0]$\\
    $i = 0$ \\
   \REPEAT
   \STATE candidate distance  = $\infty$
   \FOR{$j=0$ {\bfseries to} $L + 1$}
   \STATE Candidate configuration: $c_{j} = c_{i} + a_{j}$
   \STATE Candidate parameters: $\theta_{j} = \theta_{i,c_{j}}$
   \STATE Candidate compression rate: $\alpha_{j}$
   \STATE $\hat{g}_{j} = \hat{g}_{i} + FIT(\theta_{i}, \delta\theta_{i,j})^{\frac{1}{2}}$
   \STATE $\hat{f}_{j} = |\alpha_{j} - \alpha|FIT(\theta_{j}, \theta_{j})^{\frac{1}{2}}$
   \IF{$\hat{g}_{j} +  \lambda\hat{f}_{j} < $ candidate distance}
   \STATE $\theta_{i+1} = \theta_{j}$
   \STATE $\hat{g}_{i+1} = \hat{g}_{j}$
   \STATE candidate distance = $\hat{g}_{j} +  \lambda\hat{f}_{j}$
   \ENDIF
   \ENDFOR
   \STATE $i = i + 1$
   \UNTIL{$\alpha_{i} \leq \alpha$}
\end{algorithmic}
\end{algorithm}

\paragraph{Computational Details} As a result of the look-ahead associated with path planning, we would have to compute $L + 1$ candidate empirical Fisher traces at each iteration. Although individual computation is far faster than computing, for example, the Hessian matrix, this is still an expensive process for DNN models. Instead, we use the previous iterations' empirical Fisher traces for all actions associated with further quantization. As such, the new metric tensor is only computed for pruning steps - a direct speed increase of $L + 1$ for each step. We observe that quantization is less sensitive to this one-step delay than pruning. Layer-based quantization importance is calculated via an additional sum across all parameters within the layer, whereas global pruning masks are computed for each parameter across the network.

\section{Experiments}\label{sec:Experiments}

In the section, we apply FITCompress in a wide range of computer vision and NLP tasks. We compare FITCompress to the current state-of-the-art methods that provide their quantization configurations and pruning masks, which allows us to compute the compression metric.

\subsection{Compression Metric}
There are many possible target devices for neural network deployment, each with their own idiosyncrasies, storage, processing methods, and costs. Efficiency greatly depends on specific kernel-implementations, and inference latency or power consumption might be a deceptive indicator of compression. FITCompress is designed to be hardware-agnostic. We therefore use the percentage of Bit-Operations (BOPs) relative to the full-precision unpruned model as a proxy metric for compression. The BOPs metric generalizes floating-point operations (FLOPs) and multiply-and-accumulate operations (MACs) to heterogeneously quantized neural networks because they cannot be efficiently used to evaluate integer arithmetic operations ~\cite{freire2022computational}. This metric is commonly used in the network compression literature, and we follow its implementation from \cite{baskin2021uniq} and \cite{van2020bayesian}. 

\subsection{Experiments on Low Resolution Image Classification}

In the first set of experiments, we use LeNet-5 (4.37 GBOPs) and VGG-7 (631 GBOPs) to evaluate FITCompress on MNIST~\cite{lecun_bottou_bengio_haffner_1998} and CIFAR-10~\cite{krizhevsky2009learning}, respectively. In each experiment, we report the mean top-1 accuracy change over 3 runs. The experiment results are reported in Table~\ref{table:toy}. Experiment details are provided in Table \ref{table:training-hyperparameters} from Appendix~\ref{appendix:experimental setup}. We compare FITCompress to pruning-only methods (GAL~\cite{lin2019optimal}, AFP~\cite{ding_ding_han_tang_2018}, CFP~\cite{singh2020leveraging}, and HBFP~\cite{basha2022deep}),  an MPQ-only method (DQ~\cite{uhlich2019mixed}), and joint pruning-MPQ methods (DJPQ~\cite{wang2020differentiable} and BB~\cite{van2020bayesian}). We only consider recently published methods that provide model pruning masks and quantization configurations as these are required to compute BOPs.

We observe that FITCompress compresses LeNet-5 to 78\% of the rate reported by BB, while losing half the accuracy lost by the BB-generated model. Similarly, FITCompress compresses VGG-7 to 79\% and 48\% of the rates reported by BB and DJPQ respectively, while gaining  0.02\% accuracy. In contrast, the BB and DJPQ models lose 1.09\% and 1.51\% accuracy respectively. BB performs aggressive quantization (most LeNet-5 and VGG-7 layers in 2 bits) with very light pruning. DJPQ performs moderate quantization (most VGG-7 layers in 4 bits) with aggressive pruning in the first 3 layers. In contrast, FITCompress provides moderate quantization (LeNet-5 layers in 3-4 bits and VGG-7 layers in 3-5 bits) with 97.6\% and 94.1\% pruning on LeNet-5 and VGG-7 respectively.  FITCompress configurations are in Appendix~\ref{appendix:model configurations}. FITCompress provides a practical advantage over methods that only consider structured pruning (e.g., DJPQ and BB). See Appendix \ref{appendix: unstructured_pruning} for details.

\begin{table}[!ht]
\fontsize{7.5pt}{7.5pt}\selectfont
\centering
  \caption{Results for LeNet-5 on MNIST and VGG-7 on CIFAR-10. "-" indicates that the corresponding results are not reported in the literature. We define {\it precision} as W/A, where W and A are the bit-widths assigned to weights and activations respectively. "M" indicates heterogeneous bit-width. We include our own full-precision (FP) baseline model performance. "$\dagger$" indicates an MPQ setting restricted to power-of-two bit-widths.}
  \label{table:toy}
  \begin{tabular}{lccccc}
    \toprule
    &&
    \multicolumn{2}{c}{LeNet-5 on MNIST (FP Acc: 99.36\%)}
    & 
    \multicolumn{2}{c}{VGG-7 on CIFAR-10 (FP Acc: 93.05\%)}
    \\\cmidrule(r){3-4}\cmidrule(l){5-6}
    Method     & Precision & $\Delta$Acc. (\%) & Rel. BOPs (\%) & $\Delta$Acc. (\%) & Rel. BOPs (\%) \\
    \midrule  
    GAL & 32/32  & -0.21 & 4.40 & - & -     \\
    AFP     & 32/32 & -1.47 & 3.59 & - & -      \\
    CFP     & 32/32 & -0.86 & 2.02 & - & -  \\
    HBFP     & 32/32 & -0.57 & 2.02 & - & -  \\
    DQ$^\dagger$  & M/M & - & - & -1.46 & 0.54     \\
    DQ  & M/M  & - & - & -1.46 & 0.48     \\
    DJPQ  & M/M  & - & - & -1.51 & 0.48     \\
    DJPQ$^\dagger$  & M/M & - & -  & -1.62 & 0.46     \\
    BB     & M/M & -0.06 $\pm$ 0.03 & 0.36 $\pm$ 0.01 & -1.09 $\pm$ 0.04 & 0.29 $\pm$ 0.00  \\
    \textbf{FITCompress}     & M/M & \textbf{-0.03} $\pm$ 0.00 & \textbf{0.28} $\pm$ 0.01 & \textbf{+0.02} $\pm$ 0.06 & \textbf{0.23} $\pm$ 0.01  \\
    \textbf{FITCompress$^\dagger$}     & M/M & \textbf{-0.10} $\pm$ 0.06 & \textbf{0.44} $\pm$ 0.00 & \textbf{-1.03} $\pm$ 0.02 & \textbf{0.29} $\pm$ 0.01  \\
      \bottomrule
  \end{tabular}
\end{table}

\subsection{Experiments on Higher Resolution Image Classification}

In the second set of experiments, we use ResNet-18 (1858 GBOPs) , ResNet-34 (3752 GBOPs), and ResNet-50 (4187 GBOPs) and evaluate FITCompress on ImageNet~\cite{deng2009imagenet}. We report the mean top-1 accuracy change over 3 runs. Experiment details are provided in Table \ref{table:training-hyperparameters} from Appendix~\ref{appendix:experimental setup}. We compare FITCompress to pruning-only methods (Taylor~\cite{molchanov2019importance}, LCCL~\cite{dong2017less}, SOKS~\cite{9755967}, ASKs~\cite{LIU202132}, BCNet~\cite{su2021bcnet}, PFP-B~\cite{liebenwein2020provable}, CCEP~\cite{shang2022neural} and HBFP~\cite{basha2022deep}), MPQ-only methods (DNAS~\cite{wu2018mixed}, HAWQ-V3~\cite{yao2021hawq} and HAQ~\cite{wang2019haq}), and joint pruning-MPQ methods (VIBNet+DQ~\cite{dai2018compressing, uhlich2019mixed}, OBC~\cite{frantar2022optimal}, DJPQ~\cite{wang2020differentiable} and BB~\cite{van2020bayesian}). The results are shown in Table~\ref{table:imagenet}. We observe that FITCompress provides the best model size vs. performance trade-off even on a complex dataset like ImageNet on larger networks like variations of ResNet. Unlike BB and DJPQ, which perform aggressive quantization and light pruning, FITCompress provides more balance between pruning and quantization. FITCompress configurations are shown in Appendix~\ref{appendix:model configurations}. In Figure ~\ref{figure:ablation-path}, we illustrate how FITCompress interleaves pruning and quantization to compress ResNet-18 on ImageNet.

\begin{figure}[!ht]
\begin{center}
\centerline{\includegraphics[scale=0.4]{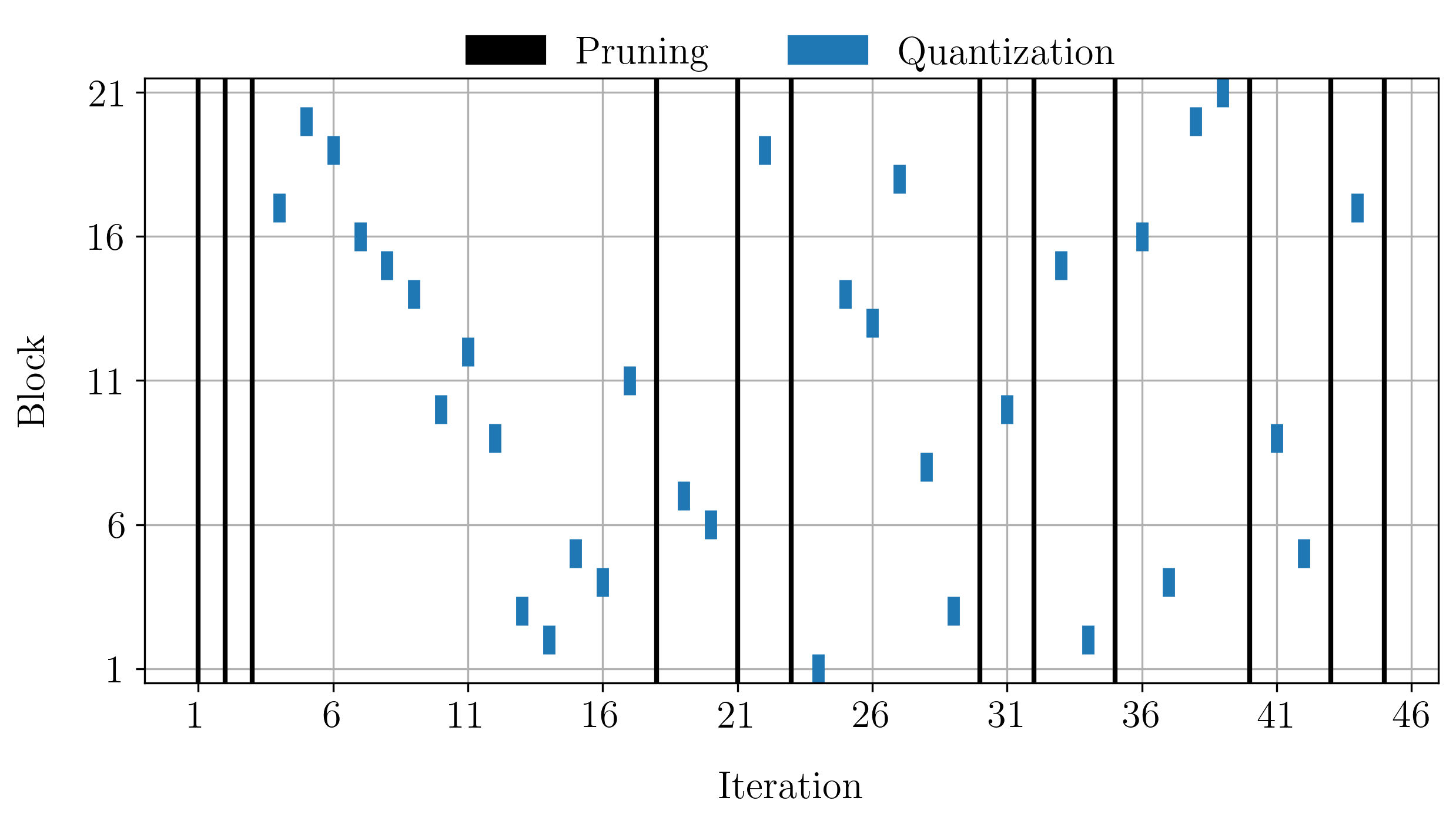}}
\caption{An example of FITCompress' compression path using ResNet-18 on ImageNet. At each iteration, FITCompress chooses whether a given layer (indexed on the vertical axis from 1-21) should be quantized or pruned (and if so by how much), until the desired model size is achieved.}
\label{figure:ablation-path}
\end{center}
\end{figure}

\begin{table}
\fontsize{7.5pt}{7.5pt}\selectfont
\centering
  \caption{Results for ResNet-18, ResNet-34, and ResNet-50 on ImageNet. We report the mean performance change over 3 runs. "-" indicates that the corresponding results are not reported in the literature. We define {\it precision} as W/A, where W and A are the bit-widths assigned to weights and activations respectively. "M" indicates MPQ. We use the pre-trained models provided by PyTorch~\cite{paszke2019pytorch} and include the full-precision (FP) baseline model performance.}
  \label{table:imagenet}
  \begin{tabular}{lccccccc}
    \toprule
    &&
    \multicolumn{2}{l}{ResNet-18 (FP Acc: 69.76\%)}
    & 
    \multicolumn{2}{l}{ResNet-34 (FP Acc: 73.31\%)}
    & \multicolumn{2}{l}{ResNet-50 (FP Acc: 76.13\%)}
    \\\cmidrule(r){3-4}\cmidrule(l){5-6}\cmidrule(l){7-8}
    Method     & Precision & $\Delta$Acc. (\%) & Rel. BOPs (\%) & $\Delta$Acc. (\%) & Rel. BOPs (\%) & $\Delta$Acc. (\%) & Rel. BOPs (\%) \\
    \midrule 
    Taylor & 32/32 & - & - & -0.48  & 75.20  & - & -     \\
    LCCL & 32/32 & -3.65 & 65.40 & -0.43  & 74.14  & - & -     \\
    ASKs & 32/32 & -0.87 & 59.34 & -0.47  & 57.34  & - & -     \\
    PFP-B  & 32/32 & -4.09 & 56.88 & - & - & - & -     \\
    SOKS  & 32/32 & -1.26 & 45.05 & -0.49 & 44.02  & - & -     \\
    BCNet  & 32/32 & - & - & - & - & -0.46 & 48.90     \\
    CCEP & 32/32 & - & - & - & - & -1.26 & 35.91   \\
    HBFP  & 32/32 & - & - & - & - & -5.69 & 25.55     \\
    HAQ  & M/M & - & - & - & - & -0.67 & 13.20     \\
    OBC  & M/M & -4.25 & 6.67 & -3.01 & 6.67 & -4.66 & 6.67     \\
    DNAS & M/M  & -1.45 & 4.73 & -0.75 & 5.26 & - & - \\
    HAWQ-V3  & M/M & -2.75 & 2.91 & - & - & -3.27 & 2.78     \\
    VIBNet+DQ & M/M & -1.22 & 2.13 & - & - & - & -     \\
    DQ & M/M & -1.34 & 2.10 & - & - & - & -     \\
    DJPQ  & M/M & -0.94 & 1.67 & - & - & - & -     \\
    BB  & M/M & -1.51 $\pm$ 0.18 & 1.38 $\pm$ 0.03 & - & - & - & -     \\
    \textbf{FITCompress}     & M/M & \textbf{-0.92} $\pm$ 0.12 & \textbf{1.38} $\pm$ 0.06 & \textbf{-0.95} $\pm$ 0.15 & \textbf{2.80} $\pm$ 0.08 & \textbf{-1.05} $\pm$ 0.23 & \textbf{2.63} $\pm$ 0.05  \\
    
      \bottomrule
  \end{tabular}
\end{table}

\subsection{Experiments on NLP Tasks}

Training a large, over-parameterized transformer and then compressing it results in better accuracy than training a smaller model directly \cite{li2020train}. In the third set of experiments, we apply FITCompress to BERT-Base~\cite{devlin2018bert} and evaluate model performance on SST-2~\cite{socher-etal-2013-recursive} and MNLI-m\cite{williams2018broadcoverage} from the GLUE benchmark~\cite{wang2018glue}, as well as SQuAD~\cite{rajpurkar2016squad}. We compare FITCompress to a pruning-only method (FPTQFT~\cite{kwon2022fast}), a MPQ-only method (Q-BERT~\cite{shen2020q}), and a joint pruning-MPQ method (OBC~\cite{frantar2022optimal}). For strong results, our training protocol includes knowledge distillation. Similarly to \cite{shen2020q}, activations and embeddings are kept at 8-bits.  Experiment details are provided in Table \ref{table:training-hyperparameters} from Appendix~\ref{appendix:experimental setup}. The results are displayed in Table~\ref{table:nlp}. We observe that FITCompress provides the best model size vs. performance trade-off. Notably, our results on NLP tasks suggest that BERT is challenging to compress, even with the combination of pruning and quantization. Complementary methods, such as the ones proposed in~\cite{mingma} could further speed up implementations and lower memory requirements.

\begin{figure}[t]
    \centering
    \includegraphics[scale=0.35]{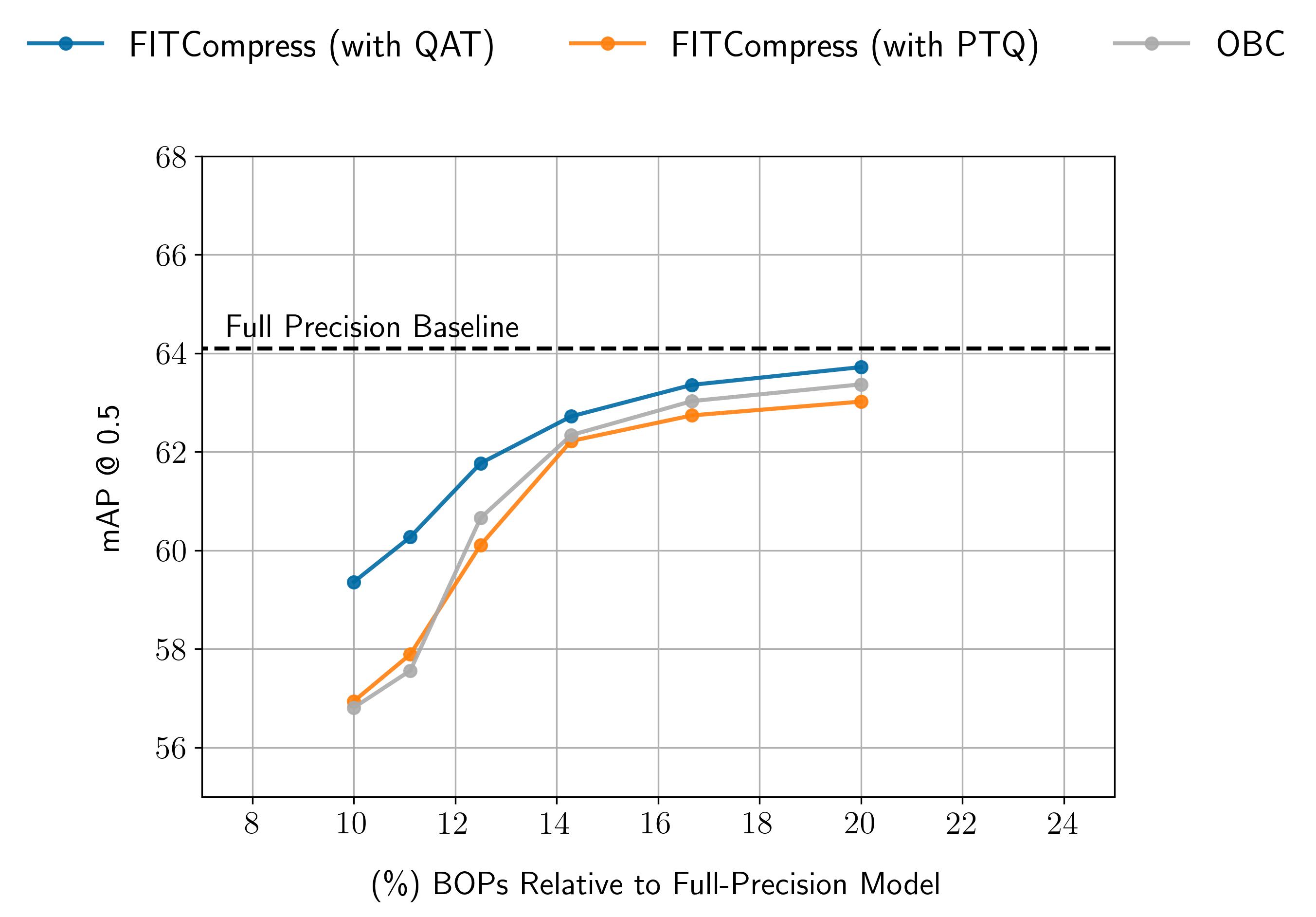}
    \caption{Comparing FITCompress and OBC across different compression constraints using YOLOv5m on COCO. We report the mean mAP at a threshold IoU of 0.5 over 3 runs.}
    \label{fig:yolov5m}
\end{figure}

\begin{table}[!ht]
\fontsize{7.5pt}{7.5pt}\selectfont
\centering
  \caption{Results for BERT on various NLP tasks. We report the mean performance change over 3 runs.}
  \label{table:nlp}
  \begin{tabular}{llcc}
    \toprule &
     Method   & $\Delta$ Performance & Rel. BOPs (\%) \\
    \midrule 
     & FPTQFT & -1.76 & 60.00 \\
     & OBC & -4.71 & 10.00 \\
     \textit {SQuAD (FP $F_1$: 88.20)} & Q-BERT & -0.33 & 3.13 \\ 
    & \textbf{FITCompress} & \textbf{0.33} $\pm$ 0.04 & \textbf{3.10} $\pm$ 0.04 \\
    \midrule 
    & FPTQFT & -1.08 & 60.00 \\
    \textit {SST-2 (FP Acc: 92.66\%)} & Q-BERT & -0.34  & 3.13 \\
    & \textbf{FITCompress} & \textbf{-0.32} $\pm$ 0.06 & \textbf{3.10} $\pm$ 0.03 \\
    \midrule 
    & FPTQFT & -2.02 & 60.00 \\
    \textit {MNLI-m (FP Acc: 85.01\%)} & Q-BERT & -0.11  & 3.13 \\
     & \textbf{FITCompress} & \textbf{-0.13} $\pm$ 0.07 & \textbf{3.10} $\pm$ 0.04 \\
    \bottomrule
      
  \end{tabular}
\end{table}

\subsection{Experiments on Object Detection}

In the final set of experiments, we use the YOLOv5m~\cite{https://doi.org/10.5281/zenodo.7347926} architecture on COCO~\cite{lin2015microsoft} to evaluate FITCompress in an object detection task. We compare FITCompress to OBC~\cite{frantar2022optimal}. To the best of our knowledge, no other works consider this learning task and architecture in a pruning or MPQ setting. For fair comparison, we extend FITCompress to a post-training quantization (PTQ) setting, that is, no QAT is performed.  Experiment details are provided in Table \ref{table:training-hyperparameters} from Appendix~\ref{appendix:experimental setup}. Results are shown in Figure~\ref{fig:yolov5m}. FITCompress (with PTQ) yields better performance than OBC under 12\% relative BOPs. In contrast, FITCompress with QAT yields superior performance for all compression constraints. 

\section{Conclusion}

In this work, we presented FITCompress, a novel algorithm that leverages the FIM and employs path planning through compression space to effectively select a combination of pruning mask and MPQ configuration. Our experimental evaluations on various computer vision and natural language processing benchmarks demonstrate that FITCompress outperforms existing state-of-the-art methods, achieving a superior compression-performance trade-off. 

\paragraph{Broader Impact and Future Work.}

FITCompress is a general model compression algorithm that can facilitates the training and deployment of efficient neural networks. On one hand, this could increase the efficiency of models deployed on edge devices (e.g. less power consumption for models deployed on servers, and longer battery lives for models running on mobile devices).  Our work could be viewed in the context of democratizing machine learning and providing access to those without industrial-scale compute. On the other hand, compression might have unpredictable effects on a given model. ~\cite{paganini2020prune} suggests that sparse models may affect individual classes instead of aggregate statistics, potentially yielding unfair models. We encourage further research in this direction for compressed and fair models. For even stronger results, one could combine FITCompress with state-of-the-art quantization implementations such as LSQ~\cite{esser2020learned} and PACT~\cite{choi2018pact}. 


\bibliographystyle{unsrt}
\bibliography{neurips_2023}
\newpage

\newpage

\appendix
\section{Further Algorithmic Details} \label{sec: further_alg_details}

\subsection{Standard Information Geometric Results}

\begin{proposition}
The FIM defines an estimate for the natural metric tensor of the statistical manifold induced by the parameterized model, obtained infinitesimally from an expansion of the KL divergence.
\end{proposition}
\begin{proof}
Here we describe the discrete case; however, the continuous follows similarly. 
\begin{align*}
    & D_{KL}(p_{\theta} || p_{\theta+\delta\theta}) = \sum_{D}{p_{\theta}\log{\frac{p_{\theta}}{p_{\theta+\delta\theta}}}} \\
    & \approx \sum_{D}{p_{\theta}{  \log{p_{\theta}} - p_{\theta}\left[ \log{p_{\theta}} + {\frac{\nabla_{\theta}p_{\theta}}{p_{\theta}}}^{T}\delta\theta + \frac{1}{2}\delta\theta^{T}\left(\frac{\nabla^{2}_{\theta}p_{\theta}}{p_{\theta}} - \frac{\nabla_{\theta}p_{\theta} \nabla_{\theta}{p_{\theta}}^{T}}{p_{\theta}^2}\right)\delta\theta \right] }} \\
    & \approx -\left[\sum_{D}\nabla_{\theta}{p_{\theta}}\right]^{T}\delta\theta - \frac{1}{2}\delta\theta^{T}\left[\sum_{D}\nabla_{\theta}^{2}{p_{\theta}}\right]\delta\theta + \frac{1}{2}\delta\theta^{T}\left[\sum_{D}p_{\theta}\nabla_{\theta}\log{p_{\theta}} {\nabla_{\theta}\log{p_{\theta}}}^{T}\right]\delta\theta \\
    & \approx -\left[\nabla_{\theta}\sum_{D}{p_{\theta}}\right]^{T}\delta\theta - \frac{1}{2}\delta\theta^{T}\left[\nabla_{\theta}^{2}\sum_{D}{p_{\theta}}\right]\delta\theta + \frac{1}{2}\delta\theta^{T}\left[\sum_{D}p_{\theta}\nabla_{\theta}\log{p_{\theta}} {\nabla_{\theta}\log{p_{\theta}}}^{T}\right]\delta\theta \\
    & \approx \frac{1}{2}\delta\theta^{T}\left[\sum_{D}p_{\theta}\nabla_{\theta}\log{p_{\theta}} {\nabla_{\theta}\log{p_{\theta}}}^{T}\right]\delta\theta \\
    & \approx \frac{1}{2}\delta\theta^{T}\left[\frac{1}{N}\sum_{D}\nabla_{\theta}\log{p_{\theta}} {\nabla_{\theta}\log{p_{\theta}}}^{T}\right]\delta\theta \\
    & = \frac{1}{2}\delta\theta^{T} \hat{I}(\theta) \delta\theta
\end{align*} 
\end{proof}
In Riemannian geometry, Rao's distance measure between two points on the statistical manifold $(\theta_1, \theta_2)$ follows from the integral of these quadratic differentials along a path joining $\theta_1$ and $\theta_2$, given by:
\begin{equation}
\left|\int_{\theta_1}^{\theta_2} [\delta\theta^{T}I(\theta)\delta\theta]^{\frac{1}{2}}\right|
\end{equation}
The geodesic (shortest path) between these two points is typically of interest. 

\subsection{The FIT Approximation}

Recent work stochastically approximates the EF \cite{zandonati2022fit}.

\begin{proposition}
FIT is obtained (under mild noise assumptions) by taking an expectation over the Fisher-Rao quadratic differential.
\end{proposition}

\begin{proof}
\begin{align*}
    \Omega &= \mathbb{E}\left[\delta\theta^{T} I(\theta) \delta\theta\right] \\
    &= \mathbb{E}[\delta\theta]^{T} I(\theta) \mathbb{E}[\delta\theta] + Tr(I(\theta)Cov[\delta\theta]) \\
\end{align*}
It is assumed that the random compression noise is symmetrically distributed around a mean of zero: $\mathbb{E}[\delta\theta] = 0$, and uncorrelated: $Cov[\delta\theta] = Diag(\mathbb{E}[\delta\theta^2])$. 
This yields the FIT heuristic:
\begin{align*}
    \Omega &= Tr\left(I(\theta)Diag(\mathbb{E}[\delta\theta^2])\right) \\
\end{align*} This is employed in an approximate form:
\begin{equation*}
    \Omega = \sum_{l}^{L} Tr(\hat{I}(\theta_{l})) \cdot  Diag(\hat{\mathbb{E}}_{l}[\delta\theta^2]).
\end{equation*}
\end{proof}
In this case, the trace of the EF admits a very simple computational form:
\begin{proposition}
The trace of the EF can be computed via the sum of the square of the gradient.
\end{proposition}

\begin{proof}
The trace is a linear operator, which allows each individual estimator to be extracted from the summation. Additionally, the trace of the second moment matrix can then be computed by the norm of the vector of parameters from which it is composed:
\begin{align*}
    Tr[\hat{I}(\theta)] =& Tr\left[{\frac{1}{N}\sum_{i=1}^{N} \nabla f(z_{i}, \theta)\nabla f(z_{i}, \theta)^{T}}\right] \\
    =& {\frac{1}{N}\sum_{i=1}^{N} Tr\left[\nabla f(z_{i}, \theta)\nabla f(z_{i}, \theta)^{T}\right]} \\
    =& {\frac{1}{N}\sum_{i=1}^{N} \nabla f(z_{i}, \theta)^{T} \nabla f(z_{i}, \theta)} \\
    =& {\frac{1}{N}\sum_{i=1}^{N} ||\nabla f(z_{i}, \theta)||^{2}} \\
\end{align*}
\end{proof}

\section{Ablation Study}
\label{appendix:ablation_studies}

\begin{figure}[h!]
    \centering
    \includegraphics[scale=0.65]{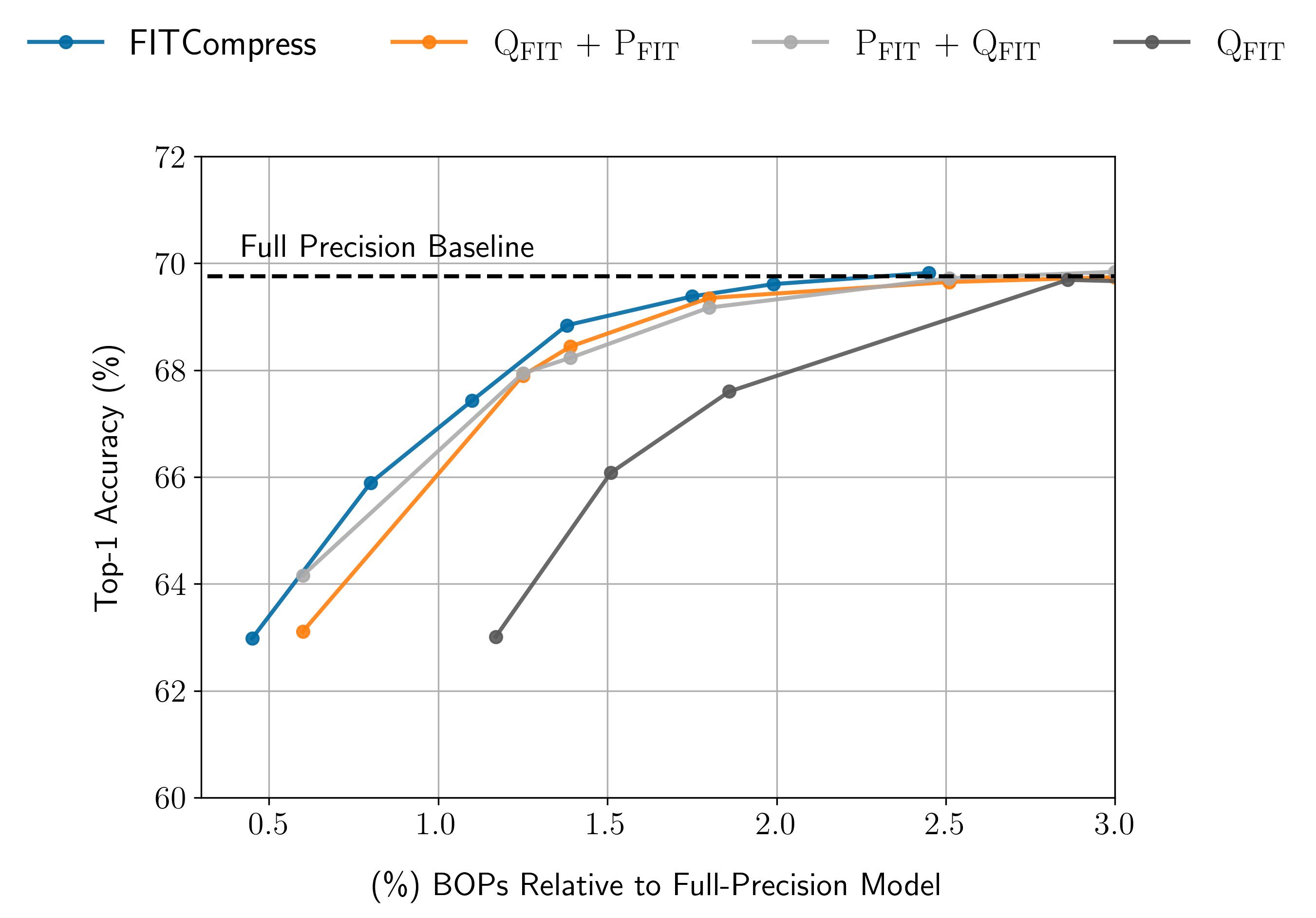}
    \caption{Comparing FITCompress with sequential applications of quantization and pruning at different compression levels. For reference we also include quantization only. We do not include pruning only. This is because we cannot simultaneously compress ResNet-18 to the 0.5-3\% relative BOPs range with pruning only and preserve top-1 accuracy above 60\% on ImageNet.}
    \label{fig:ablation_study_resnet18}
\end{figure}

In this section, we perform an ablation study to highlight the importance of path planning in FITCompress. In particular, we compare FITCompress with the quantize-then-prune ($Q_{FIT} + P_{FIT}$) and prune-then-quantize ($P_{FIT} + Q_{FIT}$) regimes. For $Q_{FIT}$, we generate several quantization configurations using the FIT sensitivity metric. This is then combined with $P_{FIT}$, where we prune with FIT as an importance heuristic using manually-selected pruning percentages to fulfill the desired compression constraint. Our ablation study repeats the findings of \cite{zhang2021training} (i.e. pruning and quantization are non-commutative). Results are displayed in Figure \ref{fig:ablation_study_resnet18}.

\clearpage

\section{Experimental Setup}
\label{appendix:experimental setup}

In this section, we detail our full experimental set-up. Additional optimization can be made in the training recipe, but this lies outside the scope of our study. We use NVIDIA A100-SXM4 40GB GPUs. We estimate that our total usage emits approximately 30 kgCO$_2$eq.

\begin{table}[!ht]
\caption{We list the hyperparameters that we used to find the optimal configuration and to train the models presented in Tables~\ref{table:toy}, \ref{table:imagenet} and \ref{table:nlp}, as well as Figure \ref{fig:yolov5m}.}
\resizebox{\textwidth}{!}{\begin{tabular}
{@{}lccccccccccccc@{}}\toprule
                    & \multicolumn{2}{c}{LeNet-5 on MNIST} & & \multicolumn{2}{c}{VGG-7 on CIFAR-10} & & \multicolumn{2}{c}{ResNet-18/34/50 on ImageNet} & & \multicolumn{2}{c}{BERT-Base on NLP Tasks} &  \multicolumn{2}{c}{YOLOv5m on COCO} \\

Optimizer           & Adam        &        &   & Adam        &        &&    SGD          &          &  &  AdamW          && SGD            \\ 
Scheduler           & Cosine Annealing LR       &       &   & Cosine Annealing LR        &        &   & Cosine Annealing LR          &         & &        Linear && Lambda LR            \\ 
Epochs     & 120             &             &   & 300            &            &   & 90               &               &  & (3, 5)                 && 300                  \\ 
Batch Size          & 128             &           &   & 128             &             &   & 128               &               &  &  32                &&    256              \\ 
Learning Rate       & $10^{-3}$             &          &   & $10^{-3}$             &            &   & $5\times 10^{-3}$              &              &  & $(2, 5)\times 10^{-5}$                && $10^{-2}$                 \\ 
Weight Decay        & $0$       &       &   & $0$       &     &   & $5\times 10^{-4}$ &         &  & 0  && $5\times 10^{-5}$           \\
Quantization schedule $Q$        & ${2, 3, 4, 5}$       &       && $${2, 3, 4, 5, 8}$$       &       &   &${2, 3, 4, 5, 8}$        &   &   &  ${2, 3, 4, 5, 8}$ && ${4, 8, 10, 12, 14, 16}$         \\
Pruning schedule $\kappa$        & $1-10^{-\frac{n}{13}}, n = 1,...,40$       & -      &   & $1-10^{-\frac{n}{13}}, n = 1,...,40$       &       &   & $1-10^{-\frac{n}{13}}, n = 1,...,40$ &         &  & $1-10^{-\frac{n}{13}}, n = 1,...,40$    && $1-10^{-\frac{n}{13}}, n = 1,...,40$            \\ \bottomrule
\end{tabular}}
\label{table:training-hyperparameters}
\end{table}

\section{FITCompress Model Configurations}

In this section, we provide detailed visualizations of the pruning and MPQ configurations obtained via FITCompress for each model discussed in Section \ref{sec:Experiments}.

\label{appendix:model configurations}
\begin{figure}[ht]
    \centering
    \includegraphics[scale=0.7]{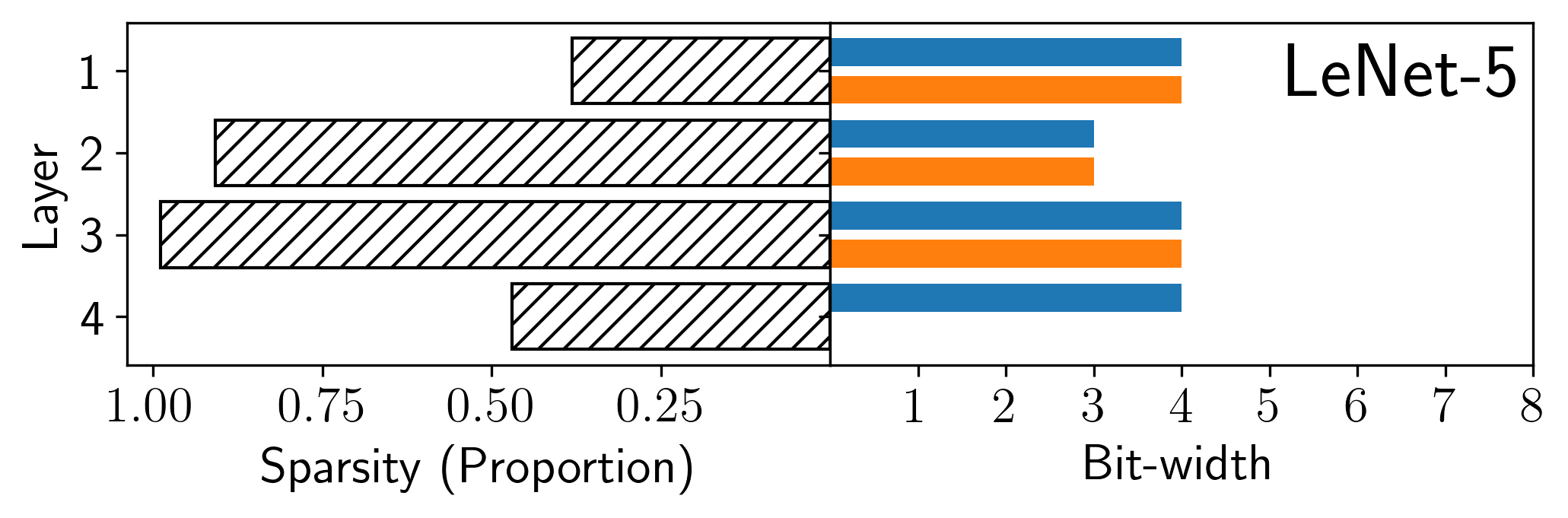}
    \includegraphics[scale=0.7]{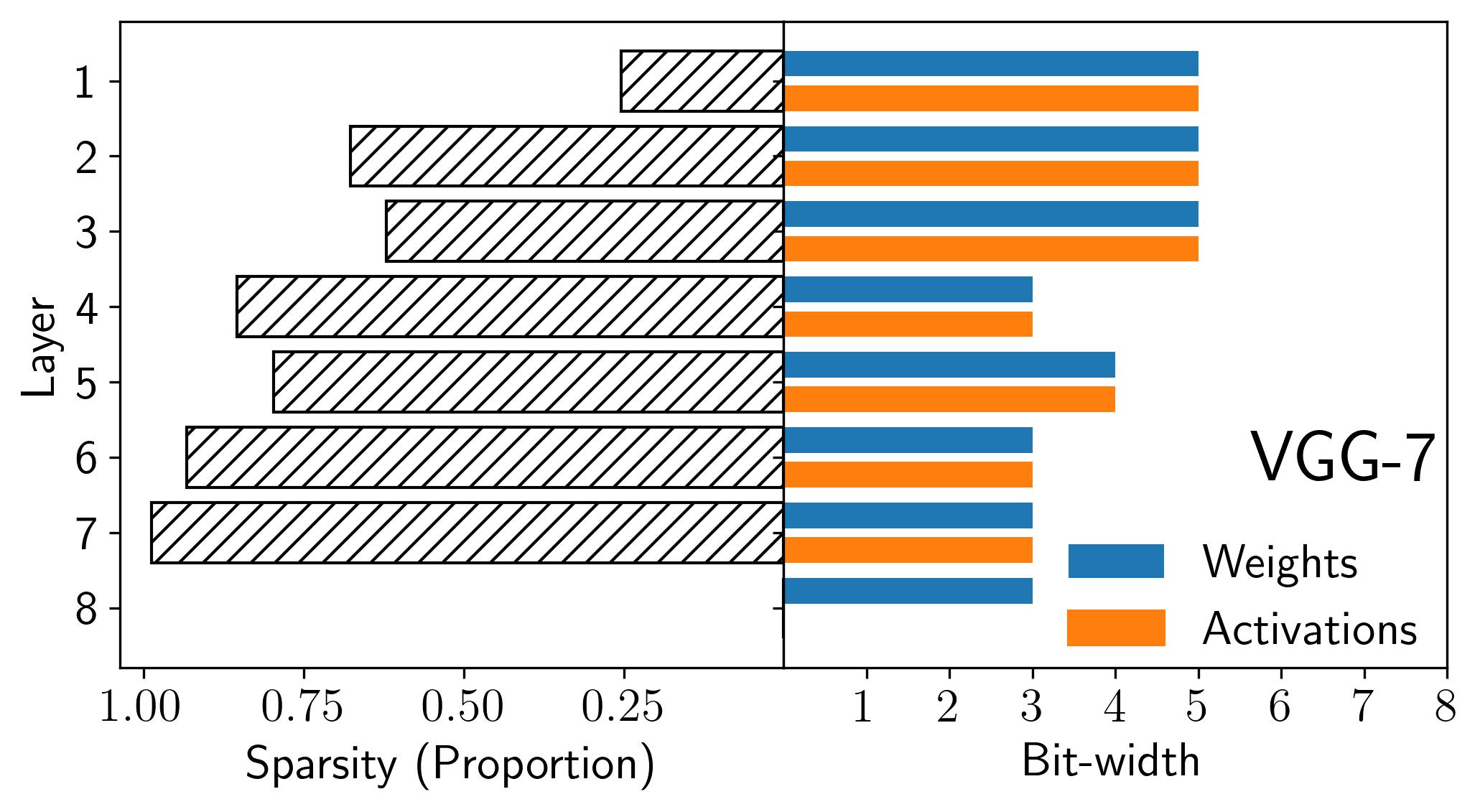}
    \caption{FITCompress-generated model configurations for LeNet-5 and VGG-7.}
    \label{fig:lenet_vgg_configurations}
\end{figure}

\begin{figure}[ht]
\centering
    \includegraphics[scale=0.45]{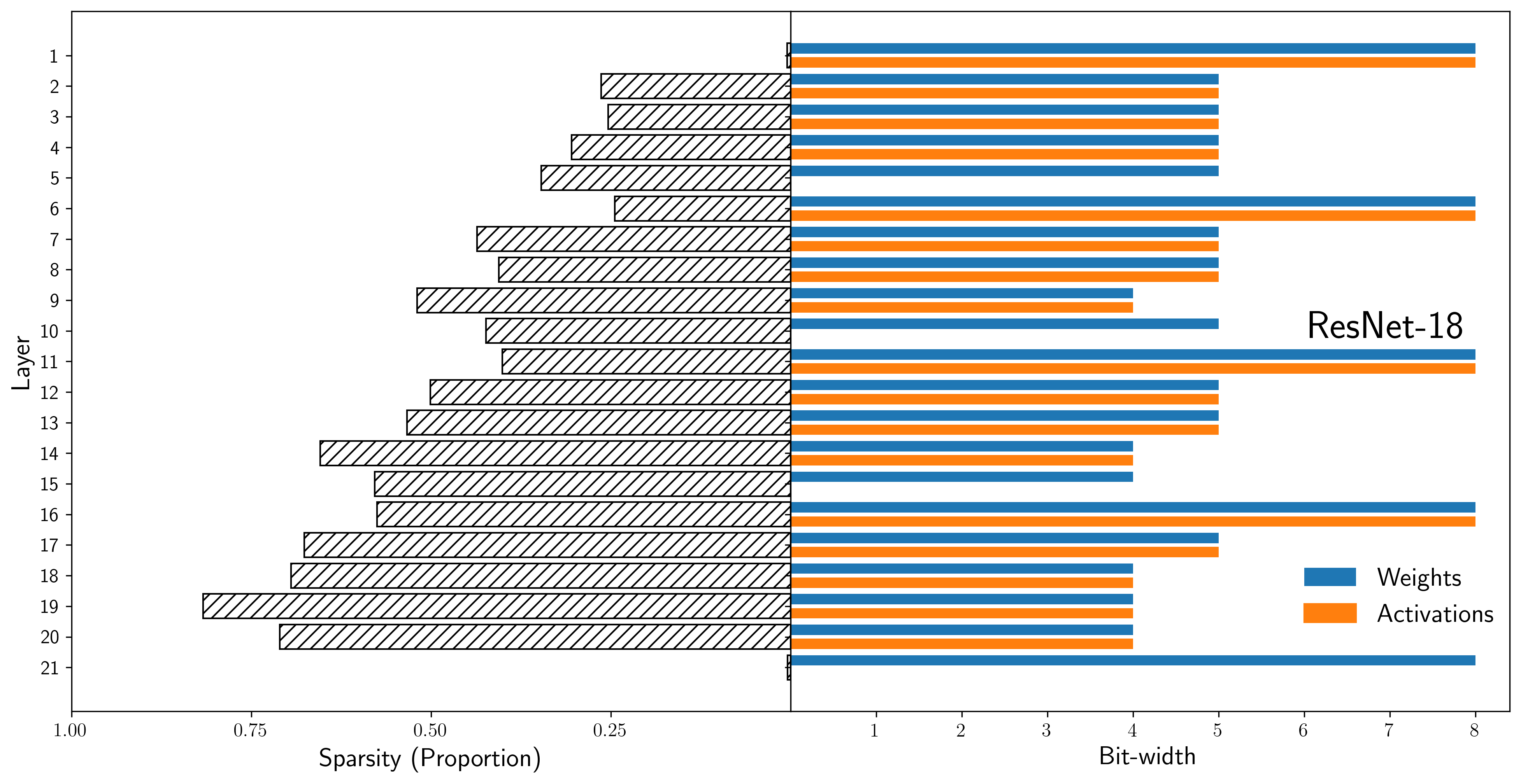}
    \caption{FITCompress-generated model configuration for ResNet-18.}
    \label{fig:resnet18_configuration}
\end{figure}

\begin{figure}[ht]
\centering
    \includegraphics[scale=0.45]{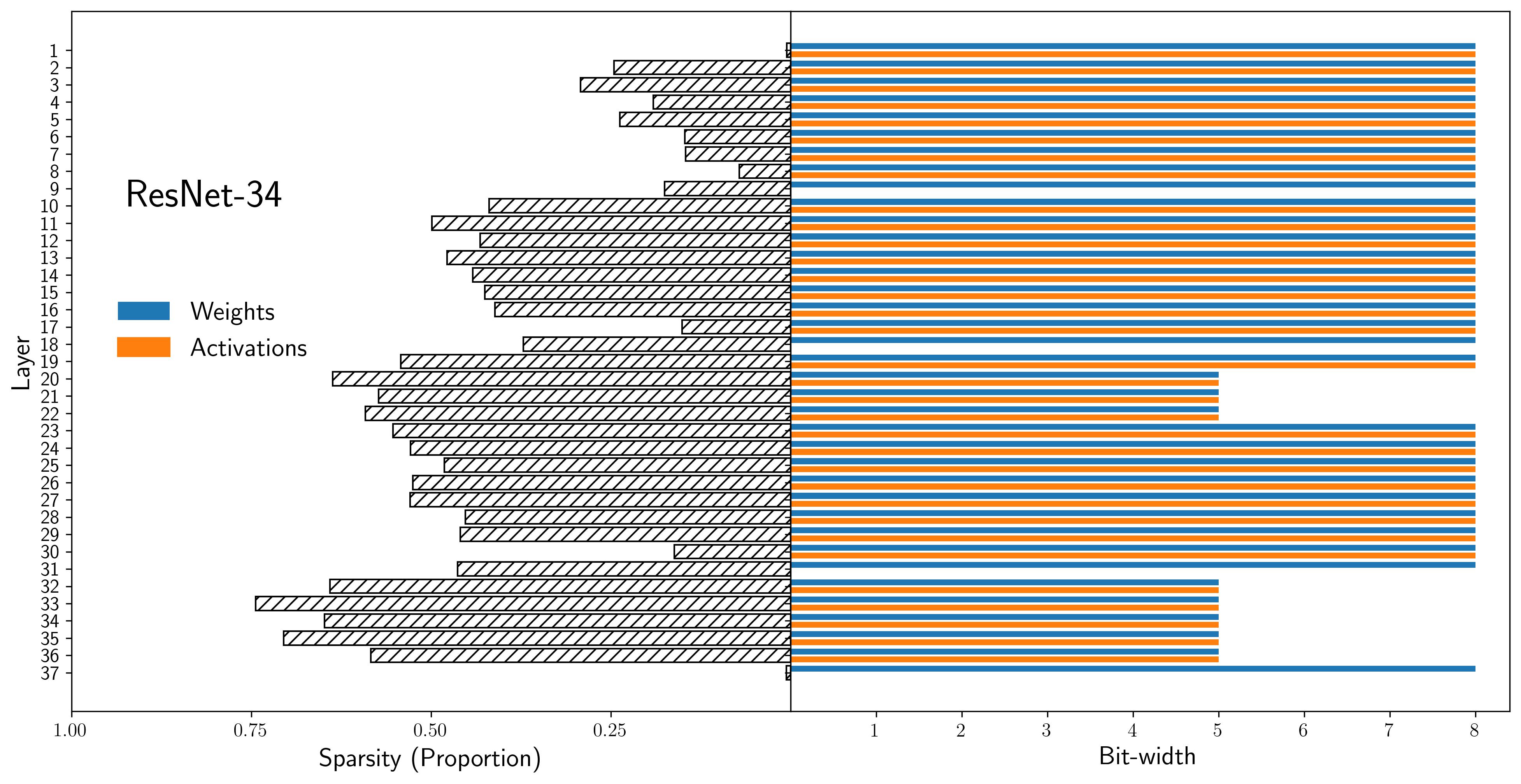}
    \caption{FITCompress-generated model configuration for ResNet-34.}
    \label{fig:resnet34_configuration}
\end{figure}

\begin{figure}[ht]
\centering
    \includegraphics[scale=0.48]{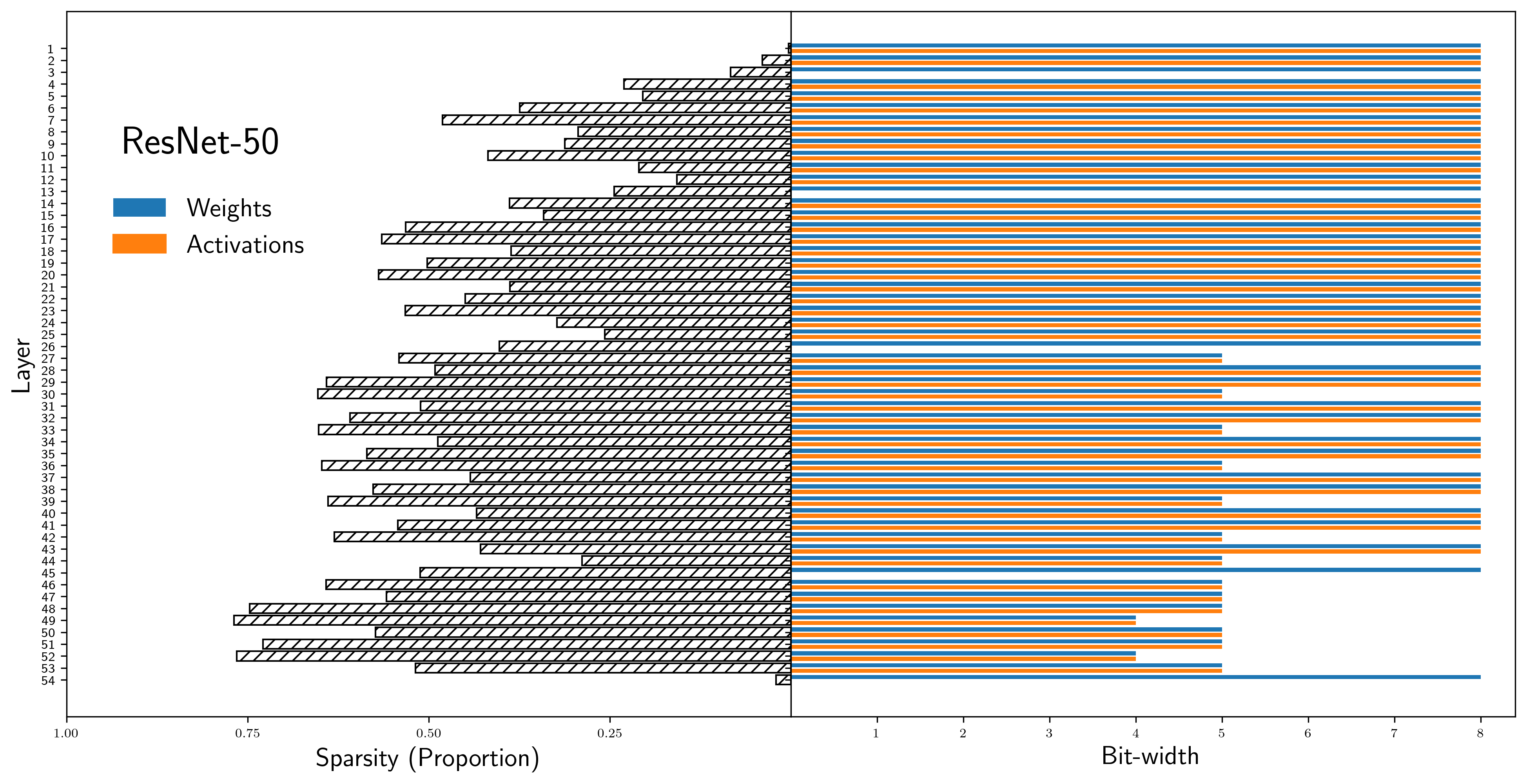}
    \caption{FITCompress-generated model configuration for ResNet-50.}
    \label{fig:resnet50_configuration}
\end{figure}
\clearpage

\section{A Comment on Unstructured Pruning} \label{appendix: unstructured_pruning}
\begin{figure}[ht]
    \centering
    \includegraphics[scale=0.8]{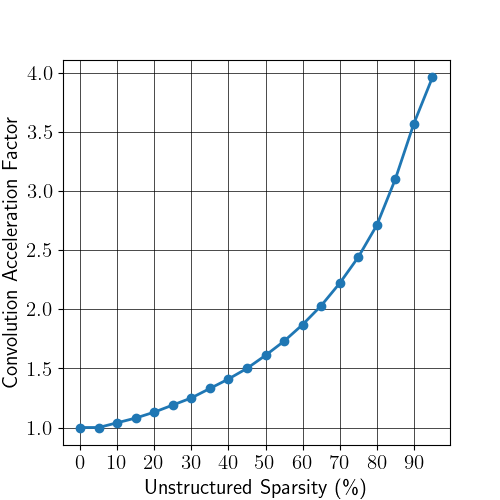}
    \caption{Acceleration of convolution vs. unstructured sparsity measured directly on a NeuPro-M processing chip. The pruning acceleration was calculated on synthesized data where zeroes were randomly distributed across weight vectors in ResNet-50.}
    \label{fig:convolution_acceleration}
\end{figure}

In contrast to methods that combine MPQ with structured pruning (e.g. BB and DJPQ), FITCompress combines MPQ with unstructured pruning. This has the advantage of yielding models with more sparsity and higher weight and activation precision compared to methods like BB and DJPQ (see Section \ref{sec:Experiments}). While structured pruning is typically favored over unstructured pruning in literature, several authors provide promising methods that are bridging the gap \cite{elsen2019fast, gale2020sparse}. Secondly, unstructured pruning has been shown to yield higher performing models compared to structured pruning methods \cite{cai2022structured, gale2019state}. We also tested unstructured sparsity on hardware to challenge the idea that it may be hardware-unfriendly. Our experiments suggest that unstructured pruning can drastically accelerate neural network operations like convolution compared to the unpruned case. We illustrate this in Figure~\ref{fig:convolution_acceleration}.

\end{document}